\documentclass{article} % For LaTeX2e
\usepackage{iclr2027_conference,times}
\iclrfinalcopy
\usepackage{amsmath,amsfonts,bm}

\def\eqref#1{equation~\ref{#1}}
\def\1{\bm{1}}

\DeclareMathAlphabet{\mathsfit}{\encodingdefault}{\sfdefault}{m}{sl}
\SetMathAlphabet{\mathsfit}{bold}{\encodingdefault}{\sfdefault}{bx}{n}

\usepackage{hyperref}
\usepackage{url}
\usepackage{booktabs}
\usepackage{multirow}
\usepackage{graphicx}
\usepackage[table]{xcolor}
\usepackage{threeparttable}

\usepackage{algorithm}
\usepackage{algpseudocode}
\usepackage{float}
\usepackage{tabularx}
\usepackage{array}
\usepackage{caption}

\title{FlashGaze: Training-Free Multi-Scale Patch Pruning for Efficient Video Understanding}

\author{
Ziye Zhu\textsuperscript{1,2} \quad
Yanghao Zhou\textsuperscript{2} \quad
Lixing Tan\textsuperscript{2} \\
\bf Jialiang Kang\textsuperscript{2} \quad
Shuxuan Li\textsuperscript{2,3} \quad
Xiao Yang\textsuperscript{2}\thanks{Corresponding author} \\
\textsuperscript{1}Peking University \quad
\textsuperscript{2}Canva Research \quad
\textsuperscript{3}Sun Yat-sen University
}
\begin{document}
\maketitle
\fancyhead{}

%========================Abstract===============================

\begin{abstract}

Multimodal Large Language Models (MLLMs) have demonstrated strong performance in video understanding, yet efficiently processing long, high-resolution videos remains challenging. Such videos often contain substantial spatiotemporal redundancy, and processing redundant visual tokens can incur avoidable computational overhead.
Many existing methods prune visual tokens during or after vision transformer (ViT) encoding, leaving much of the encoding cost unaddressed. Some approaches prune patches before encoding but rely on learned auxiliary networks for patch selection, incurring additional training and inference overhead.
To address these limitations, we propose FlashGaze, a training-free method that reduces spatiotemporal redundancy before ViT encoding without introducing auxiliary networks. FlashGaze uses pixel-space differences as a proxy for information loss and employs Quadtree Dynamic Programming to jointly optimize patch dropping, merging, and keeping under a fixed budget.
Experiments on two MLLM backbones across multiple benchmarks demonstrate substantial efficiency gains while largely preserving accuracy. On Qwen3-VL-8B, FlashGaze retains \textbf{98\%} of the full-input baseline accuracy on LongVideoBench while achieving up to \textbf{5.4$\times$} and \textbf{17$\times$} speedups in ViT encoding and MLLM prefill, respectively, and reducing peak GPU memory usage by a factor of \textbf{1.8}. These efficiency gains enable the model to process videos with more frames and higher resolutions on the same GPU hardware, unlocking video understanding at scales previously out of reach.

\end{abstract}

%========================Intro=================================

\section{Introduction}
\label{sec:intro}

Recent advances in multimodal large language models (MLLMs) have demonstrated strong capabilities in video understanding~\citep{qian2024streaming,bai2025qwen3,zhang2024llava}, with growing attention to long-form and high-resolution scenarios~\citep{mangalam2023egoschema,wu2024longvideobench,zhou2025mlvu,fu2025video,hu2026video}. Emerging applications such as embodied intelligence and autonomous driving further demand accurate, low-latency processing of such videos. However, long, high-resolution videos produce lengthy visual token sequences that incur substantial computational and memory overhead and may exceed the context limits of MLLMs, hindering their deployment in real-world applications.

MLLMs typically process all visual tokens in a video, despite substantial temporal redundancy from static content across frames and spatial redundancy from large low-information regions. These redundant tokens impose considerable unnecessary computation. Existing methods therefore seek to reduce visual tokens while preserving informative content. For example, ToMe~\citep{bolya2022token} reduces tokens within the ViT, VisionZip~\citep{yang2025visionzip} compresses tokens after the ViT, and FastV~\citep{chen2024image} prunes tokens inside the MLLM.

\begin{figure*}[!t]
  \centering
  \includegraphics[width=\textwidth]{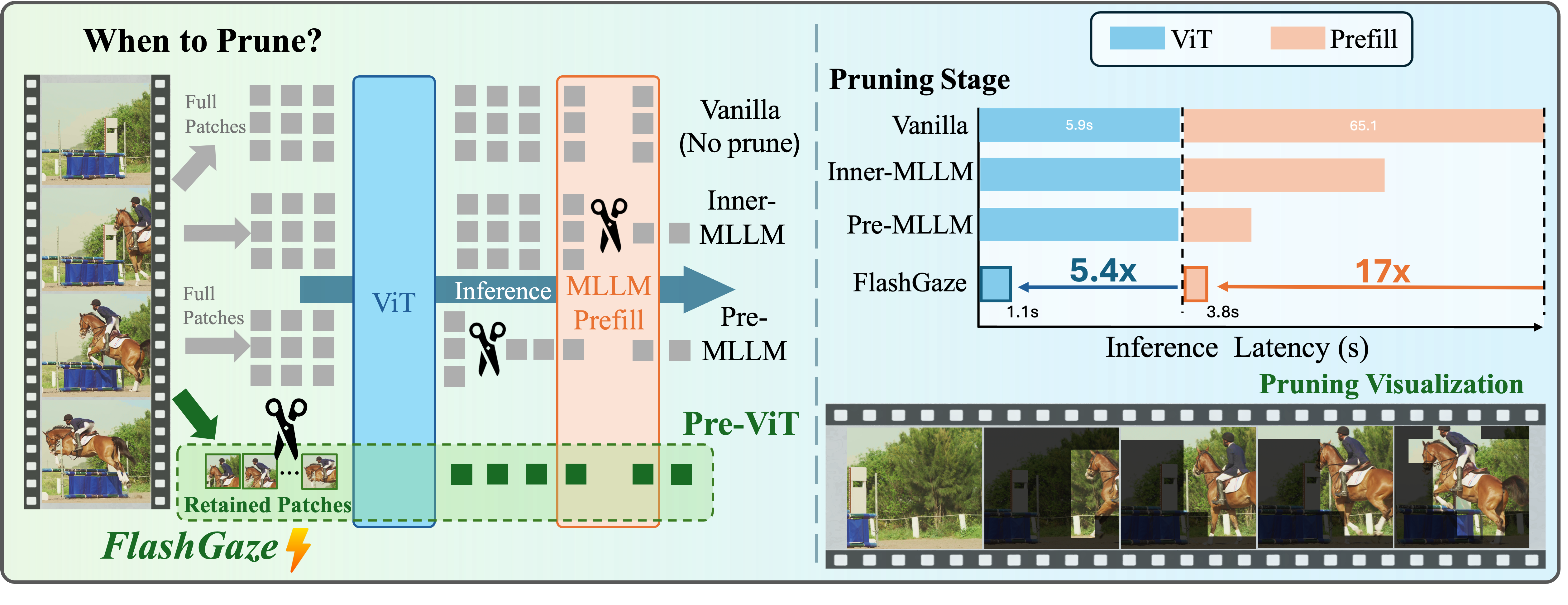}
    \caption{We propose FlashGaze, a training-free method that enables MLLMs to process 1.5K-frame and 4K-resolution videos. Existing methods prune visual tokens before or inside the MLLM, leaving substantial redundant computation in the ViT. In contrast, FlashGaze prunes patches directly at the video input stage, retaining \textbf{98\%} of the full-input baseline accuracy while speeding up the ViT and prefill stages by up to \textbf{5.4$\times$} and \textbf{17$\times$}, respectively. We also visualize the pruning results, showing FlashGaze selectively retains informative visual patches.}

  \label{fig:1}
\end{figure*}

Although these methods have shown promising results, as illustrated in Fig.~\ref{fig:1}, they prune visual tokens only before or inside the MLLM. Consequently, part or all of the ViT, and even several MLLM layers in some cases, still process the full set of visual tokens. The later token pruning is applied, the more unnecessary computation redundant tokens incur, limiting the ability of MLLMs to scale to longer and higher-resolution videos. Another line of work explores token pruning at the video input stage. AutoGaze~\citep{shi2026attend}, for example, employs an autoregressive network to select visual patches, removing redundant ones before the ViT. However, such methods typically introduce additional network components and may rely on specially constructed training data, resulting in non-negligible training and inference overhead. In contrast, we observe that temporal and spatial redundancy can be assessed directly in pixel space, as static regions exhibit minimal pixel variation across frames and low-information areas display high local similarity, enabling effective patch selection before the ViT without a learned selector. 

\begin{figure}[htb]
    \centering
    \includegraphics[width=\linewidth]{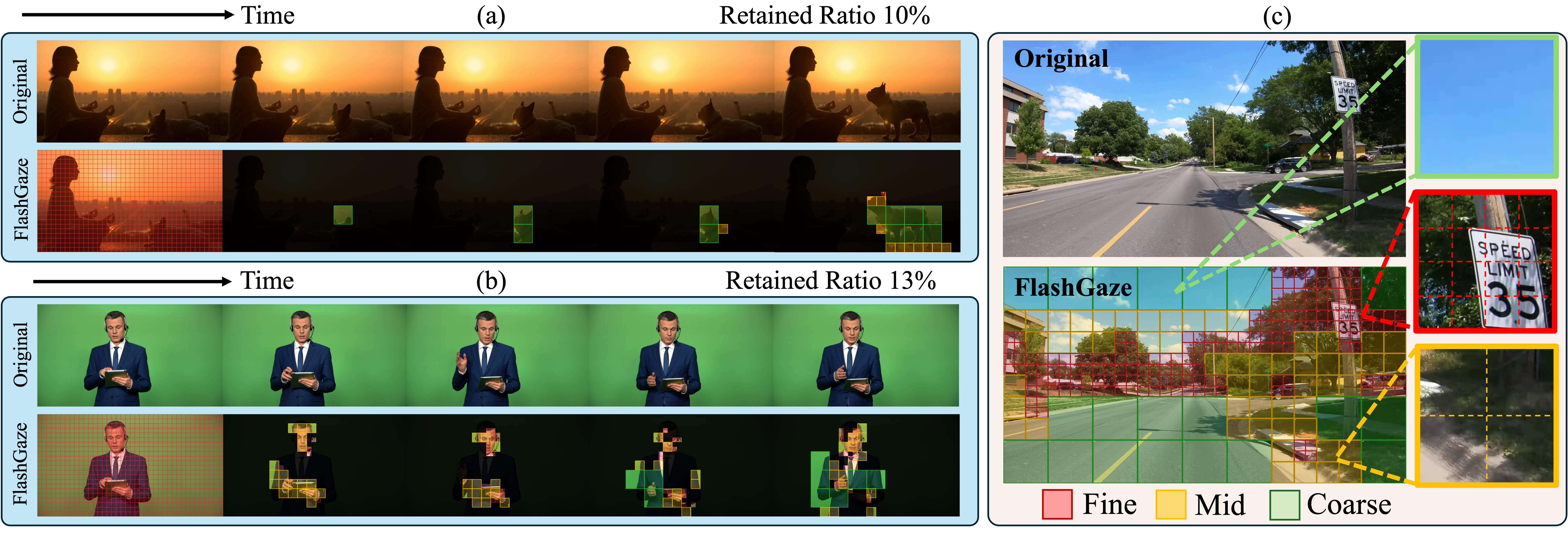}
    \caption{\textbf{Visualization of pruned videos.} FlashGaze mainly retains patches in regions that change over time in (a) and (b). In the continuously changing driving scene in (c), FlashGaze uses coarse-scale patches for low-information areas such as the sky, while preserving fine-scale patches for detail-rich regions such as traffic signs.}
    \label{fig:2}
\end{figure}

Based on this observation, we propose FlashGaze, a training-free method that prunes visual patches directly at the video input stage. For each patch, FlashGaze considers three actions: dropping it, merging it with neighboring patches, or retaining it at its original resolution. FlashGaze estimates the cost of each action in pixel space: the dropping cost is measured by the difference between the current patch and its corresponding historical patch, while the merging cost is measured by the difference before and after merging. A quadtree naturally captures the hierarchical structure of multi-scale merging, and dynamic programming (DP) efficiently solves the selection problem under a given patch budget. FlashGaze thus employs Quadtree DP to determine the action assignment, dropping temporally redundant patches while allocating fine-scale patches to detail-rich areas and coarse-scale patches to low-information areas (Fig.~\ref{fig:2}). 

We evaluate FlashGaze on two representative MLLMs, Qwen3-VL-8B and NVILA-8B-Video, across a range of video understanding benchmarks. Notably, FlashGaze retains 98\% of the full-input baseline accuracy while accelerating the ViT and prefill stages by up to $5.4\times$ and $17\times$, respectively. 

Our major contributions are as follows:

\begin{itemize}
    \item We propose FlashGaze, a training-free method that performs patch pruning at the video input stage via Quadtree DP, guided by pixel-space dropping and merging costs, to simultaneously reduce temporal and spatial redundancy.

    \item Extensive experiments on two representative MLLMs and multiple video understanding benchmarks demonstrate the effectiveness and efficiency of FlashGaze, while unlocking longer visual contexts for improved video understanding.

    \item Our analysis reveals that FlashGaze largely preserves visual representations, enables MLLMs to allocate more attention to dynamic visual tokens, and assigns fine-scale patches to information-rich regions, thereby maintaining high accuracy.
\end{itemize}

\section{Related Work}
\subsection{Long-Form Video Understanding}
Long-form video understanding requires integrating information over extended temporal horizons~\citep{mangalam2023egoschema,wu2024longvideobench,chen2026scaling,xie2026fluxmem,rege2026agentic}, yet processing longer videos substantially increases the number of visual tokens~\citep{chen2026scaling}, raising computational and memory costs and potentially exceeding the context capacity of MLLMs. Prior work addresses this by extending context capacity~\citep{chen2025longvila}, maintaining compact memories~\citep{song2024moviechat,he2024ma}, or selectively retrieving relevant visual evidence~\citep{wang2024videoagent}. These methods improve how long-range information is represented or accessed, but efficiently processing the underlying visual input remains a challenge when dense sampling introduces substantial spatiotemporal redundancy.

\subsection{Visual Token Pruning and Compression}
Visual token reduction can occur at several stages. DynamicViT and ToMe prune or merge tokens within vision encoders~\citep{rao2021dynamicvit,bolya2022token}; VisionZip compresses encoder outputs~\citep{yang2025visionzip}, while FastV prunes tokens within the language model~\citep{chen2024image}. These approaches reduce downstream computation, leaving part or all of the encoding cost unaddressed. Video methods further exploit temporal redundancy: LongVU, VideoChat-Flash, and FastVID compress visual tokens after encoding~\citep{shen2024longvu,li2026videochat,shen2025fastvid}, while TimeChat-Online removes inter-frame redundancy~\citep{yao2025timechat}. Addressing encoding cost directly, STC reuses encoder features~\citep{wang2026accelerating} and EarlyTom applies compression early within the encoder~\citep{wang2026earlytom}. Selection before encoding avoids processing redundant content altogether: PS3 selectively encodes high-resolution regions~\citep{shi2025scaling}, PixelPrune is limited to static document and GUI images~\citep{wang2026pixelprune}, and AutoGaze learns to select multi-scale patches autoregressively~\citep{shi2026attend}. In contrast, FlashGaze jointly reduces temporal and spatial redundancy with pixel-space costs and Quadtree DP, without training or auxiliary networks

%========================Methods==============================
\begin{figure*}[!t]
    \centering
    \includegraphics[width=\linewidth]{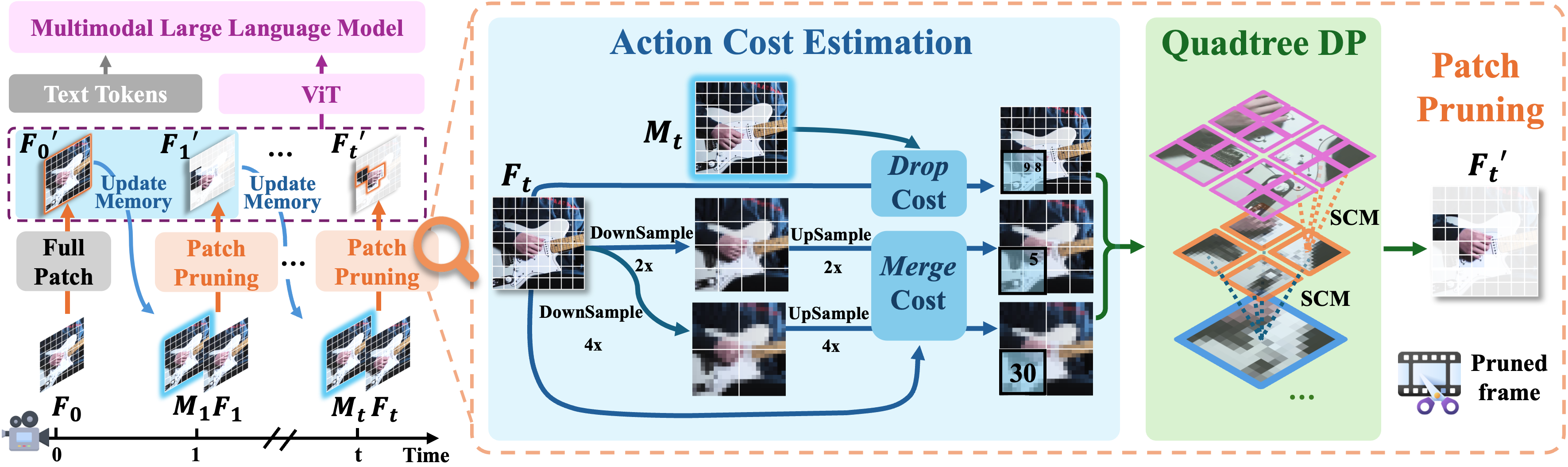}
    \caption{\textbf{Processing pipeline of FlashGaze.}  FlashGaze performs patch pruning on video frames sequentially, based on visual memory and current frames (left). During patch pruning, FlashGaze estimates the costs of $\mathrm{Drop}$ and $\mathrm{Merge}$ using current frame $F_t$ and memory $M_t$ (middle). Given the costs and budget, FlashGaze determines the action assignment via Quadtree DP (right).}
    \label{fig:3}
\end{figure*}

\section{Method}
\subsection{Problem Formulation}

Scaling MLLMs to long, high-resolution videos calls for reducing redundant computation not only in the LLM backbone, but also in the vision encoder. We therefore target spatiotemporal redundancy at the video input stage. The central challenge is to decide which regions require new visual information and at what spatial resolution, without relying on encoded features or a learned selector. We formulate this challenge as budget-constrained patch selection. Given a video frame $F_t$ at time $t$ and a patch budget $\mathcal{B}$, we seek a compressed representation $F'_t$ that minimizes information loss while retaining at most $\mathcal{B}$ patches:
\begin{equation}
\label{eq:opti}
F_t'^{*} =
\arg\min_{F_t'} \mathcal{L}(F_t',F_t)
\quad
\mathrm{s.t.}\quad
N_t \leq \mathcal{B},
\end{equation}
where $F_t'^{*}$ denotes a compressed representation obtained under this objective, $N_t$ the number of retained patches in $F_t'$, and $\mathcal{L}$ the information discrepancy between the compressed representation and the original frame.

\subsection{FlashGaze}
\label{sec:FlashGaze}

To instantiate this formulation, FlashGaze constructs $F'_t$ through a joint assignment of three patch-level actions: $\mathrm{Drop}$, $\mathrm{Merge}$, and $\mathrm{Keep}$. $\mathrm{Drop}$ omits temporally redundant patches, $\mathrm{Merge}$ represents neighboring patches at a coarser spatial scale, and $\mathrm{Keep}$ preserves patches at their original resolution. We represent the action assignment for frame $F_t$ as a set $\mathcal{A}_t$ of patch-action pairs $(p, a)$.

We use the sum of pixel-space action costs as a measure of the information discrepancy in Eq. ~\ref{eq:opti}. We further introduce a patch-count penalty $\lambda N_t$ to favor more compact representations without using the full budget $\mathcal{B}$. This yields the following regularized objective:
\begin{equation}
\mathcal{A}_t^{*}
=
\arg\min_{\mathcal{A}_t}
\left[
\sum_{(p,a)\in\mathcal{A}_t}
\operatorname{Cost}(p,a)
+
\lambda N_t
\right],
\quad
\mathrm{s.t.}\quad N_t\leq\mathcal{B}.
\label{eq:2}
\end{equation}
Here, $N_t$ denotes the number of patches retained under $\mathcal{A}_t$, and $\lambda$ controls the trade-off between information preservation and the number of retained patches.

As shown in Fig.~\ref{fig:3}, FlashGaze performs patch pruning on video frames sequentially. To provide historical information, FlashGaze maintains a visual memory $M_t$ at each time step. At each spatial location, $M_t$ stores the most recently retained patch from previously processed frames.
The first frame $F_0$ is fully retained, with $F'_0=F_0$, and initializes the visual memory $M_1$. This initialization frame is exempt from the patch budget. For each subsequent frame $F_t$ ($t\geq1$), FlashGaze uses $M_t$ as a reference to prune $F_t$ and obtain $F'_t$.
After obtaining $F'_t$, its retained patches update corresponding locations in the memory, producing $M_{t+1}$ for the next frame.

%---------------3.3--------------------------
\subsection{Patch Pruning}

During patch pruning, FlashGaze estimates the costs of different actions for each patch based on the current frame $F_t$ and the visual memory $M_t$. Given these action costs and the budget, FlashGaze solves Eq.~\ref{eq:2} via Quadtree DP to determine an action assignment.

\paragraph{Action Cost Estimation}
\label{para:cost estimation}

To reduce temporal redundancy, we estimate the dropping cost of each patch using the historical visual information stored in the visual memory. For a basic patch $p_{\mathrm{f}}$, we compare its content in the current frame $F_t$ with the corresponding content in the visual memory $M_t$ to obtain $\operatorname{Cost}(p_{\mathrm{f}}, \mathrm{Drop})$:
\begin{equation}
\operatorname{Cost}(p_{\mathrm{f}}, \mathrm{Drop}) = D_{\mathrm{pix}}( F_t(p_{\mathrm{f}}),M_t(p_{\mathrm{f}})),
\end{equation}

where $D_{\mathrm{pix}}$ denotes the pixel-wise difference between two aligned regions of the same size. To mitigate the effects of slight camera motion, we divide each region into small blocks and search a local neighborhood in the reference image for the displacement that minimizes the mean pixel-wise difference within each block. We then align the reference region accordingly and compute the patch-level difference. $\operatorname{Cost}(p_{\mathrm{f}}, \mathrm{Drop})$ reflects the difference between the current patch and the historical patch, where a higher value indicates a larger difference.

To reduce spatial redundancy, we evaluate the information loss caused by merging neighboring patches. We consider two spatial merge scales, $s\in\{\mathrm{m},\mathrm{c}\}$, where smaller and larger groups of neighboring fine-scale patches are merged into $p_{\mathrm{m}}$ and $p_{\mathrm{c}}$, respectively. For each scale $s$, we spatially downsample $F_t$ and then upsample the result to the original resolution, obtaining the reconstructed frame $\widehat{F}_t^{(s)}$. Similar to the dropping cost, the merging cost is computed as follows:
\begin{equation}
\operatorname{Cost}(p_s,\mathrm{Merge})
=
D_{\mathrm{pix}}\!\left(
F_t(p_s),\widehat{F}_t^{(s)}(p_s)
\right),
\quad
s\in\{\mathrm{m},\mathrm{c}\},
\end{equation}

where $\operatorname{Cost}(p_{s}, \mathrm{Merge})$ reflects the information difference before and after merging, scaled by the number of fine-scale patches covered by $p_s$ to ensure cross-scale comparability. Keeping a fine-scale patch unchanged incurs zero cost, i.e., $\operatorname{Cost}(p_{\mathrm{f}},\mathrm{Keep})=0$.

\paragraph{Quadtree DP}
\label{para:quadtree}

Candidate patches at different scales may overlap, so they cannot be selected independently. We organize them into a three level quadtree with fine, mid, and coarse scales. Each mid-scale node $p_{\mathrm{m}}$ contains four fine-scale patches $p_{\mathrm{f}}$, and each coarse-scale node $p_{\mathrm{c}}$ contains four mid-scale nodes. Quadtree DP computes the minimum-cost arrays from fine to mid to coarse scales, and determines the action for each region under the budget.

We denote $T_{\mathrm{f}}(p_{\mathrm{f}}, n)$, $T_{\mathrm{m}}(p_{\mathrm{m}}, n)$, and $T_{\mathrm{c}}(p_{\mathrm{c}}, n)$ as the minimum costs of representing regions covered by $p_{\mathrm{f}}$, $p_{\mathrm{m}}$, and $p_{\mathrm{c}}$ with exactly $n$ retained patches. For $s\in\{\mathrm{f},\mathrm{m},\mathrm{c}\}$,
$T_s(p_s,\cdot)$ denotes the entire cost array for the corresponding region. The local patch count $n$ is distinct from the frame-level patch count $N_t$. For a leaf node $p_{\mathrm{f}}$, the patch can either be dropped or kept unchanged. Therefore, $n\in\{0,1\}$, and the corresponding costs are as follows:
\begin{equation}
T_{\mathrm{f}}(p_{\mathrm{f}}, n)= 
\begin{cases}
\operatorname{Cost}(p_{\mathrm{f}}, \mathrm{Drop}), & n=0,\\
0, & n=1.
\end{cases}
\end{equation}

We use \textbf{Subtree Cost Merging (SCM)} to combine the cost arrays of child nodes. For a mid-scale node $p_{\mathrm{m}}$ with four fine-scale children $p_0,\ldots,p_3$, SCM enumerates all allocations $(n_0,\ldots,n_3)\in\{0,1\}^4$ and records the minimum sum of child costs for each total retained patch count $n=\sum_{i=0}^{3}n_i$, yielding $\widetilde{T}_{\mathrm{m}}(p_{\mathrm{m}},n)$. Algorithm~\ref{alg:SCM} illustrates this fine-to-mid aggregation. At higher levels, SCM uses the corresponding feasible child patch counts.

\begin{algorithm}[htb]
\caption{Subtree Cost Merging (SCM)}
\label{alg:SCM}
\begin{algorithmic}[1]
\Require Child cost arrays $T_{\mathrm{f}}(p_i,\cdot)$, $i=0,\ldots,3$, for the four children of $p_{\mathrm{m}}$
\Ensure Combined cost array $\widetilde{T}_{\mathrm{m}}(p_{\mathrm{m}},\cdot)$ for $n=0,\ldots,4$

\For{$n=0,\ldots,4$}
    \State $\widetilde T_{\mathrm{m}}(p_{\mathrm{m}},n)=+\infty$
\EndFor
\ForAll{$(n_0,n_1,n_2,n_3)\in\{0,1\}^{4}$}
    \State $n=0,\quad T_{\mathrm{total}}=0$
    \For{$i=0,\ldots,3$}
        \State $n=n+n_i,\quad
        T_{\mathrm{total}}=T_{\mathrm{total}}+T_{\mathrm{f}}(p_i,n_i)$
    \EndFor
    \State $\widetilde T_{\mathrm{m}}(p_{\mathrm{m}},n)=
    \min\bigl(\widetilde T_{\mathrm{m}}(p_{\mathrm{m}},n),T_{\mathrm{total}}\bigr)$
\EndFor
\end{algorithmic}
\end{algorithm}

We then account for the option of merging the entire mid-scale region into a single patch:
\begin{equation}
T_{\mathrm{m}}(p_{\mathrm{m}},n)=
\begin{cases}
\widetilde T_{\mathrm{m}}(p_{\mathrm{m}},n), & n\neq1,\\
\min\bigl(\widetilde T_{\mathrm{m}}(p_{\mathrm{m}},1),\operatorname{Cost}(p_{\mathrm{m}},\mathrm{Merge})\bigr), & n=1.
\end{cases}
\end{equation}

For $n=1$, DP compares the cost of retaining one fine-scale patch and dropping the other three with the cost of merging the entire region into a single mid-scale patch.

For each coarse-scale node $p_{\mathrm{c}}$, we apply the same procedure to its four mid-scale children, including the comparison with $\operatorname{Cost}(p_{\mathrm{c}},\mathrm{Merge})$ at $n=1$, to obtain $T_{\mathrm{c}}$. We then recursively combine the coarse-scale cost arrays using SCM to obtain the frame-level cost array. Finally, following Eq.~\ref{eq:2}, we select the retained patch count that minimizes the total action cost plus $\lambda N_t$ subject to $N_t\leq\mathcal{B}$, and backtrack the recorded choices to obtain the final action assignment. See Appendix~\ref{app:a} for details.

\paragraph{Integration with MLLMs}

The resulting action assignment determines which patches are sent to the ViT and at what scale. A fine-scale patch $p_{\mathrm{f}}$ keeps its original content, while $p_{\mathrm{m}}$ and $p_{\mathrm{c}}$ represent the corresponding area after 2$\times$ and 4$\times$ spatial downsampling. FlashGaze assigns each retained patch a positional embedding based on the position of its center. Then, selected patches are encoded by the ViT and fed into MLLM as usual. See Appendix~\ref{app:integ} for details.
%=====================Experiments=======================

\section{Experiments}
\subsection{Experimental Setup}
\label{subsect:Exp_setup}
\paragraph{Benchmarks}
We evaluate FlashGaze on six video understanding benchmarks spanning general, long-video, and high-resolution settings: VideoMME~\citep{fu2025video}, NExT-QA~\citep{xiao2021next}, MLVU~\citep{zhou2025mlvu}, VideoMMMU~\citep{hu2026video}, LongVideoBench~\citep{wu2024longvideobench}, and HLVid~\citep{shi2026attend}. Among them, we use the last three for main comparisons. In addition, we compare efficiency under different retained ratios on two of these benchmarks. Across all six benchmarks, we further evaluate FlashGaze with larger visual inputs across diverse tasks.

\paragraph{Comparison Methods}
We compare FlashGaze with five token compression methods: ToMe~\citep{bolya2022token}, FastV~\citep{chen2024image}, VisionZip~\citep{yang2025visionzip}, FlashVID~\citep{fan2026flashvid}, and AutoGaze~\citep{shi2026attend}, in terms of accuracy and TTFT. On Qwen3-VL-8B, we further compare accuracy, peak GPU memory, and TTFT under different retained ratios.

\paragraph{Implementation Details}
We conduct all experiments on NVIDIA H200 GPUs with Qwen3-VL-8B~\citep{bai2025qwen3} and NVILA-8B-Video~\citep{liu2025nvila} backbones. For a fair comparison, we define the retained ratio as the fraction of visual tokens remaining after the final pruning stage. TTFT includes the latency of pre-ViT video processing, pruning, ViT encoding, and MLLM prefill.

\newcommand{\vcell}[1]{%
  \multicolumn{1}{c!{\hspace{-1pt}\vrule width 0.3pt\hspace{1pt}}}{#1}}
\newcolumntype{C}{>{\centering\arraybackslash}X}

\begin{table*}[htb]
\caption{\textbf{Performance comparison with other methods.} FlashGaze achieves the highest accuracy and lowest TTFT across high-resolution and long-video benchmarks.\textsuperscript{*}FastV is omitted for NVILA-8B-Video due to pre-pruning context overflow.}
\centering
\small
\setlength{\tabcolsep}{2pt}
\renewcommand{\arraystretch}{1.15}
\begin{tabularx}{\textwidth}{@{}l *{11}{C}@{}}
\toprule
\multirow[b]{2}{*}{Method} &
\multicolumn{4}{c}{LongVideoBench} &
\multicolumn{5}{c}{VideoMMMU} &
\multicolumn{2}{c}{HLVid} 
\\
\cmidrule(lr){2-5}
\cmidrule(lr){6-10}
\cmidrule(lr){11-12}
& Perc. & Relat. & Overall & TTFT(s)
& Perc. & Comp. & Adapt. & Overall & TTFT(s)
& Overall & TTFT(s)
\\

\midrule
\rowcolor{gray!12}
\multicolumn{12}{c}{\textit{Qwen3-VL-8B}} \\
%===================50%keep====================
+ToMe
& \textbf{73.2} & 60.4 & 66.4
& \vcell{24.3}
& 78.0 & 69.0 & 48.3 & 65.1 
& \vcell{23.7}
& 71.6
& 25.7 \\

+FastV
& 72.8 & 60.4 & 66.2
& \vcell{28.7}
& 79.6 & 67.0 & \textbf{50.7} & 65.8 
& \vcell{28.2}
& 70.9
& 30.1 \\

+VisionZip
& 72.6 & 60.3 & 66.0
& \vcell{25.4}
& 79.0 & 68.3 & 49.0 & 65.4 & \vcell{24.6}
& 71.2
& 27.5 \\

+FlashVID
& 72.0 & 59.7 & 65.4
& \vcell{26.5}
& 77.3 & 64.7 & 49.3 & 63.8 & \vcell{25.8}
& 72.3
& 28.9 \\

+AutoGaze
& 72.0 & 60.3 & 65.8
& \vcell{34.9}
& 77.7 & 69.7 & 47.7 & 65.0& \vcell{37.2}
& 69.8
& 34.6 \\

\rowcolor{cyan!10}
\textbf{+FlashGaze}
& 72.3 & \textbf{61.8} & \textbf{66.7}
& \vcell{\textbf{20.7}}
& \textbf{80.9} & \textbf{71.3} & 48.7 & \textbf{67.0} & \vcell{\textbf{20.4}}
& \textbf{73.2}
& \textbf{20.5} \\

\midrule

%====================10%keep=============================
\rowcolor{gray!12}
\multicolumn{12}{c}{\textit{NVILA-8B-Video\textsuperscript{*}}} \\
+ToMe
& 61.1 & 48.0 & 54.2
& \vcell{30.7}
& 46.7 & 34.7 & 41.0 & 40.8 & \vcell{29.4}
& 49.6
& 58.0 \\

+VisionZip
& 61.3 & 47.3 & 53.9
& \vcell{34.9}
& 43.3 & 34.0 & 41.3 & 39.6 & \vcell{32.8}
& 49.3
& 65.4 \\

+FlashVID
& 61.4 & 48.9 & 54.8
& \vcell{38.9}
& 45.3 & 37.3 & 38.0 & 40.2 & \vcell{42.4}
& 52.6
& 77.4 \\

+AutoGaze
& 65.4 & \textbf{53.4} & 58.8
& \vcell{17.5}
& \textbf{53.0}& 36.3 & \textbf{42.7} & 44.0 & \vcell{15.9}
&  49.3
& 32.2 \\

\rowcolor{cyan!10}
\textbf{+FlashGaze}
& \textbf{65.8} & 53.0 & \textbf{59.0}
& \vcell{\textbf{13.7}}
& 52.3 & \textbf{38.0} & \textbf{42.7} & \textbf{44.3} & \vcell{\textbf{13.6}}
& \textbf{53.0}
& \textbf{27.3} \\

\bottomrule
\end{tabularx}
\label{tab:1}
\end{table*}

\begin{table*}[htb]
\centering
\caption{\textbf{Scalability across various benchmarks.} FlashGaze scales Qwen3-VL-8B to 1.5K-frame and 4K-resolution videos, unlocking longer context for better video understanding.}
\label{tab:2}
\scriptsize
\setlength{\tabcolsep}{4pt}
\renewcommand{\arraystretch}{0.95}
\resizebox{\textwidth}{!}{%
\begin{tabular}{@{}l*{8}{c}@{}}
\toprule
& &
& \multicolumn{2}{c}{\textit{general video}}
& \multicolumn{3}{c}{\textit{long video}}
& \textit{high-res \& long} \\

\cmidrule(lr){4-5}
\cmidrule(lr){6-8}
\cmidrule(lr){9-9}
\textbf{Models}
& \textbf{\begin{tabular}[c]{@{}c@{}}Max\\\#F\end{tabular}}
& \textbf{\begin{tabular}[c]{@{}c@{}}Max\\Res\end{tabular}}
& \textbf{\begin{tabular}[c]{@{}c@{}}VideoMME\\(w/o sub)\end{tabular}}
& \textbf{\begin{tabular}[c]{@{}c@{}}NExT-QA\\(mc)\end{tabular}}
& \textbf{\begin{tabular}[c]{@{}c@{}}VideoMMMU\\(overall)\end{tabular}}
& \textbf{\begin{tabular}[c]{@{}c@{}}LongVideoBench\\(overall)\end{tabular}}
& \textbf{\begin{tabular}[c]{@{}c@{}}MLVU\\(m-avg)\end{tabular}}
& \textbf{\begin{tabular}[c]{@{}c@{}}HLVid\\(test)\end{tabular}} \\
\midrule
LLaVA-OV-8B ~\citep{li2025llava}
& 32
& 384
& 58.2
& 79.4
& -
& 56.5
& 64.7
& 41.1\\
LLaVA-Video-7B ~\citep{zhang2024llava}
& 64
& 384
& 63.3
& 83.2
& 36.1
& 58.2
& 70.8
& - \\
VideoLLaMA3-7B ~\citep{zhang2025videollama}
& 180
& 384
& 66.2
& 84.5
& 47.0
& 59.8
& 73.0
& 38.8\\
InternVL3.5-8B ~\citep{wang2025internvl3}
& 64
& 448
& 66.0
& -
& -
& 62.1
& 70.2
& 39.9\\
VideoChat-Flash ~\citep{li2026videochat}
& 10000
& 448
& 65.3
& -
& 41.7
& 64.7
& 74.7
& 46.6\\
LongVILA-7B ~\citep{chen2025longvila}
& 2048
& 384
& 60.1
& 80.7
& -
& 57.1
& -
& 41.4\\
NVILA-8B-Video ~\citep{liu2025nvila}
& 256
& 448
& 64.2
& 82.2
& -
& 57.7
& 70.1
& 42.5\\
Qwen2.5-VL-7B ~\citep{bai2025qwen25vltechnicalreport}
& 48
& 896
& 65.1
& -
& 47.4
& 56.0
& 70.2
& 48.1\\
\midrule
Qwen3-VL-8B ~\citep{bai2025qwen3}
& 768
& 1792
& 71.5
& 83.6
& 65.3
& 66.7
& 77.3
& 71.6 \\
\rowcolor{cyan!10}
~+ \textbf{FlashGaze}
& 1536
& 3840
& \textbf{71.8}
& \textbf{84.7}
& \textbf{67.0}
& \textbf{66.9}
& \textbf{77.6}
& \textbf{73.2}  \\

\bottomrule
\end{tabular}%
}
\end{table*}

\subsection{Main comparisons}

\paragraph{Performance Comparison with Other Methods}

We compare FlashGaze with five methods on Qwen3-VL-8B and NVILA-8B-Video, using retained ratios of 50\% and 10\%, respectively. For LongVideoBench and VideoMMMU, the maximum number of sampled frames is set to 768. For HLVid, we evaluate high resolution performance using 2K-resolution videos. Under the same backbone, we use the same retained ratio across methods and compare their accuracy and TTFT latency. As shown in Table~\ref{tab:1}, FlashGaze achieves the highest accuracy and the lowest TTFT on both backbones for high resolution and long video understanding.

\paragraph{Scalability across various benchmarks}

As shown in Table~\ref{tab:2}, FlashGaze scales Qwen3-VL-8B to 1.5K-frame and 4K-resolution videos. With richer visual inputs, FlashGaze outperforms the vanilla model on all six benchmarks and achieves the highest accuracy. These results demonstrate that FlashGaze enables MLLMs to scale to longer and higher resolution videos, leading to better video understanding.

\paragraph{Efficiency under Different Retained Ratios}

To further evaluate the efficiency of FlashGaze, we compare it with three methods under different retained ratios on two benchmarks. 
As shown in Table~\ref{tab:3}, FlashGaze achieves the highest accuracy in seven settings while achieving the lowest TTFT and peak GPU memory in all settings. At a 20\% retained ratio, FlashGaze still preserves 98\% of the full input accuracy on LongVideoBench, while reducing peak GPU memory by 1.8$\times$. It also achieves a 13$\times$ TTFT speedup over the vanilla model, including 5.4$\times$ and 17$\times$ speedups in ViT encoding and MLLM prefill. See Appendix~\ref{app:b} for details. Compared with other pruning methods, FlashGaze reduces both TTFT and peak GPU memory usage.

\begin{table*}[htb]
\caption{\textbf{Efficiency under different retained ratios.} FlashGaze achieves the lowest TTFT and peak GPU memory across all settings while largely preserving accuracy.}
\centering
\small
\renewcommand{\arraystretch}{1.0}
\resizebox{\textwidth}{!}{%
\begin{tabular}{@{}c@{\hspace{8pt}}c@{}}
\setlength{\tabcolsep}{3pt}
% 左表
\begin{tabular}{lcccc}
\toprule
\multirow{2}{*}{Method} &
Retained &
L-VidBench &
Peak GPU &
\multirow{2}{*}{TTFT (s)}
\\
& Ratio & Acc. & Mem. (GiB) \\
\midrule
Qwen3-VL-8B & 100\% & 66.7 & 106.4 & 71.7\\
\midrule
%=================LVB-50%keep====================
+ToMe      & \multirow{4}{*}{50\%} & 66.4 & 82.5 & 24.3\\
+FlashVID  &                       & 65.4 & 101.2 & 26.5\\
+AutoGaze  &                       & 65.8 & 81.0 & 34.9\\
\textbf{+FlashGaze}&&\textbf{66.7}&
\textbf{80.6}&
\textbf{20.7}

\\
\midrule
%=================LVB-40%keep====================
+ToMe      & \multirow{4}{*}{40\%} & 66.6 & 82.5 & 18.6 \\
+FlashVID  &                       & 64.9 & 101.2& 20.7 \\
+AutoGaze  &                       & 65.7 & 74.0 & 25.3 \\
\textbf{+FlashGaze}&&\textbf{66.7}
&\textbf{73.1}
&\textbf{14.5} \\
\midrule
%=================LVB-30%keep====================
+ToMe      & \multirow{4}{*}{30\%} & 65.9 & 82.5 & 13.8\\
+FlashVID  &                       & 64.4 & 101.2 & 16.0\\
+AutoGaze  &                       & 65.1 & 65.5 & 18.2\\
\textbf{+FlashGaze}&&\textbf{66.3}
&\textbf{65.2}
&\textbf{9.5}
\\
\midrule
%=================LVB-20%keep====================
+ToMe      & \multirow{4}{*}{20\%} & 65.1 & 82.5 & 10.3 \\
+FlashVID  &                       & 63.6 & 101.2 & 12.2 \\
+AutoGaze  &                       & 64.1 & 59.0 & 12.1 \\
\textbf{+FlashGaze}&& \textbf{65.5} 
&\textbf{58.2}
&\textbf{5.4}
\\
\bottomrule
\end{tabular}
&
% 右表
\begin{tabular}{lcccc}
\toprule
\multirow{2}{*}{Method} &
Retained &
HLVid &
Peak GPU &
\multirow{2}{*}{TTFT (s)}
\\
& Ratio & Acc. & Mem. (GiB) \\
\midrule
Qwen3-VL-8B & 100\% & 71.6 & 105.6 & 72.3\\
\midrule
%=================HLVID-50%keep====================
+ToMe      & \multirow{4}{*}{50\%} & 71.6 & 82.3 &25.7\\
+FlashVID  &                       & 72.3 & 101 &28.9\\
+AutoGaze  &                       & 69.8 & 80.2&34.6\\
\textbf{+FlashGaze}&&\textbf{73.2}&\textbf{80.0}&\textbf{20.5}
\\
\midrule
%=================HLVID-40%keep====================
+ToMe      & \multirow{4}{*}{40\%} & 71.2 & 82.3 &20.0\\
+FlashVID  &                       & \textbf{71.3}&101 &23.0 \\
+AutoGaze  &                       & 68.2 & 72.9 & 25.4\\
\textbf{+FlashGaze}&&\textbf{71.3}&\textbf{72.4}&\textbf{14.3}
\\
\midrule
%=================HLVID-30%keep====================
+ToMe      & \multirow{4}{*}{30\%} & 71.6 & 82.5 &15.4\\
+FlashVID  &                       & 71.6 & 101.2 &18.2\\
+AutoGaze  &                       & 66.7 & 65.5&19.0\\
\textbf{+FlashGaze}&&\textbf{71.7}&\textbf{65.2}&\textbf{9.1}
\\
\midrule
%=================HLVID-20%keep====================
+ToMe      & \multirow{4}{*}{20\%} & 67.9 & 82.3 & 12.1\\
+FlashVID  &                       & \textbf{70.9} & 101.0 & 14.8\\
+AutoGaze  &                       & 54.9 & 58.4 & 13.5\\
\textbf{+FlashGaze}&& 68.2&\textbf{58.0}&\textbf{5.3}
\\
\bottomrule
\end{tabular}
\end{tabular}%
}
\label{tab:3}
\end{table*}

\begin{figure}[htb]
    \centering
    \includegraphics[width=\linewidth]{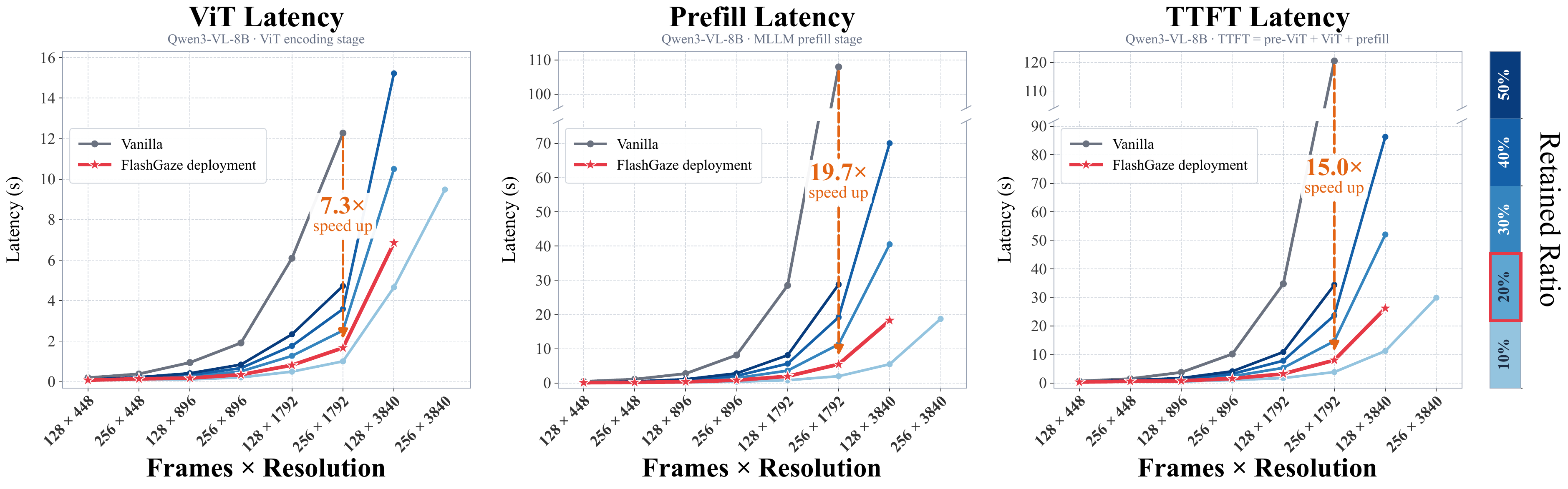}
    \caption{\textbf{Latency across input scales and retained ratios.} FlashGaze can choose different retained ratios to balance accuracy and efficiency. At a 20\% retained ratio, FlashGaze still maintains strong accuracy. At this ratio, with 256 frames and 1792 resolution, it achieves 7.3$\times$ ViT, 19.7$\times$ prefill, and 15$\times$ TTFT speedups. }
    \label{fig:4}
\end{figure}

\subsection{Ablation Studies}

\paragraph{Input Scale and Retained Ratio.}
By varying the patch budget $\mathcal{B}$ and penalty coefficient $\lambda$, we evaluate the efficiency of FlashGaze on Qwen3-VL-8B across different retained ratios, frame counts, and input resolutions. As shown in Fig.~\ref{fig:4}, Qwen3-VL-8B quickly reaches its context limit as the number of frames and input resolution increase, supporting up to 256 frames at a resolution of 1792 in our setting. At this input scale, with $\lambda=4\times10^{-3}$, FlashGaze reaches a 20\% retained ratio at 256 frames and 1792 resolution, achieving 7.3$\times$ and 19.7$\times$ speedups in ViT encoding and MLLM prefill, and a 15$\times$ TTFT speedup.

\paragraph{Cost Estimation Method}
We compare several cost estimation methods to study their impact on downstream accuracy. Besides pixel differences, we use ViT Patch Embedding~\citep{dosovitskiy2020image},
SigLIP~\citep{zhai2023sigmoid}, and LPIPS~\citep{zhang2018unreasonable} to measure
feature, semantic, and perceptual differences, respectively. Under a 30\% retained ratio with 768 sampled frames, pixel differences achieve the best scores on both HLVid and LongVideoBench, reaching 71.7 and 66.3. The other methods perform slightly worse while introducing additional feature extraction and higher inference latency. Overall, pixel differences provide a better tradeoff between accuracy and efficiency.

\paragraph{Optimization Method.}
We compare four methods for solving Eq.~\ref{eq:2} and report their downstream accuracy and cumulative action cost, $\sum \operatorname{Cost}$. As shown in Table~\ref{tab:ablation}, Quadtree DP achieves both the lowest cost and the highest accuracy, at a 30\% retained ratio. Notably, among the other three algorithms, lower cumulative cost is consistently associated with higher accuracy. This trend is consistent with our objective of reducing visual information loss by minimizing action costs.

\begin{table*}[htb]
\caption{\textbf{Ablation on cost estimation and optimization methods.}}
\centering
\small
\setlength{\tabcolsep}{2pt}
\renewcommand{\arraystretch}{1.15}
% ==================== Left Table ====================
\begin{minipage}[t]{0.49\textwidth}
\centering

\begin{tabular}{@{}lccc@{}}
\toprule
\multirow[c]{2}{*}{Method} &
\multirow[c]{2}{*}{Type} &
HLVid &
L-VidBench \\
& & Acc. & Acc. \\
\midrule
Patch embedding
& Feature-space
& 69.3
& 65.8
\\
SigLIP
& Semantic
& 65.1
& 64.2
\\
LPIPS
& Perception
& 69.0
& 66.0
\\
\textbf{Pixel (ours)}
& Pixel
& \textbf{71.7}
& \textbf{66.3}
\\
\bottomrule
\end{tabular}
\end{minipage}%
\hfill
% ==================== Right Table ====================
\begin{minipage}[t]{0.49\textwidth}
\centering

\begin{tabular}{@{}lccccc@{}}
\toprule

\multirow[c]{2}{*}{Method} &
\multicolumn{3}{c}{Candidates} &
$\sum \operatorname{Cost}$ &
HLVid \\
\cline{2-4}
& fine & mid & coarse & ($\times10^{3}$) & Acc. \\
\midrule

Random
& $\checkmark$
&
&
& 16.50
& 60.8
\\

Single-scale Greedy
& $\checkmark$
&
&
& 7.5
& 69.3
\\

Multi-scale Greedy
& $\checkmark$
& $\checkmark$
& $\checkmark$
& 3.3
& 70.6
\\

\textbf{Quadtree DP (ours)}
& $\checkmark$
& $\checkmark$
& $\checkmark$
& \textbf{1.96}
& \textbf{71.7}
\\

\bottomrule
\end{tabular}
\end{minipage}
\label{tab:ablation}
\end{table*}

\begin{figure}[htb]
    \centering
    \includegraphics[width=\linewidth]{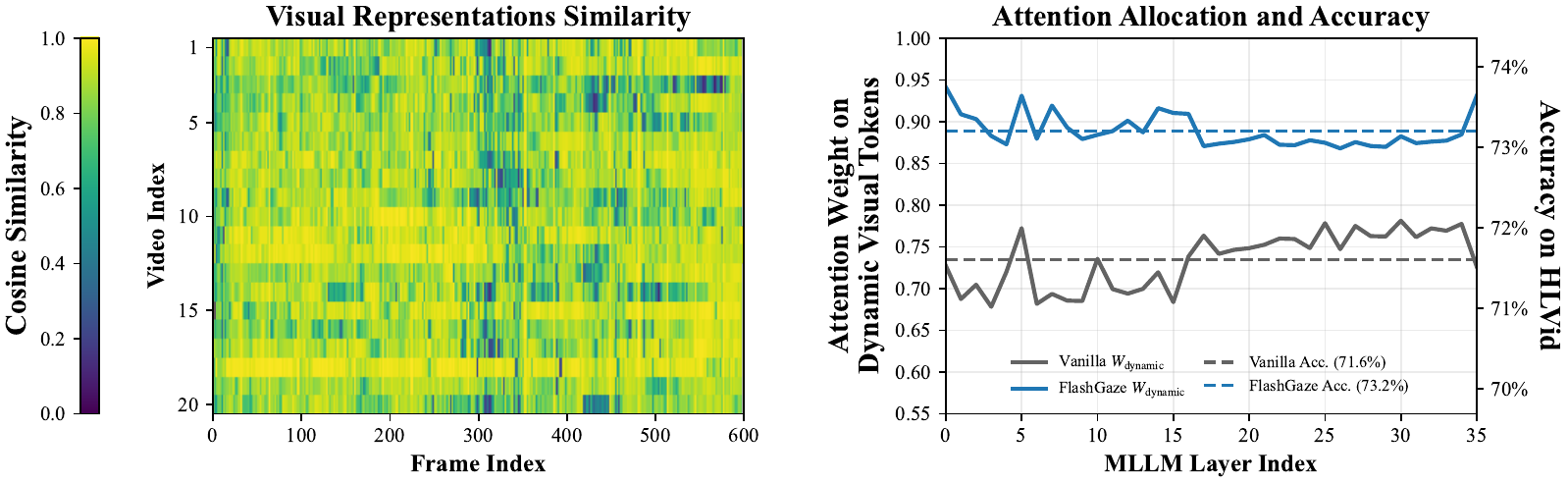}
    \caption{\textbf{Analysis of how FlashGaze enables strong video understanding.} Visual representations remain highly similar between FlashGaze and the vanilla setting (left).  FlashGaze enables MLLMs to allocate more attention to dynamic tokens and achieves better accuracy, suggesting that increased attention to dynamic tokens may contribute to the accuracy improvement (right).}
    \label{fig:5}
\end{figure}

\subsection{Analysis and Visualization}

\paragraph{Preserving Visual Representations.}
To evaluate whether pruning preserves visual representations, we compare attention weighted representations from FlashGaze and the full input on 20 HLVid videos. To handle different token counts in corresponding regions, we use the last prefill token as the query and aggregate first layer visual value vectors by their attention weights. We then compute their cosine similarity. See Appendix~\ref{app:c} for details. As shown in Fig.~\ref{fig:5} (left), the visual representations remain highly similar between FlashGaze and the vanilla setting.

\paragraph{More Attention to Dynamic Visual Tokens.}
To analyze attention to dynamic visual tokens, we identify regions that change across frames and mark their corresponding tokens as dynamic. On HLVid with a 50\% retained ratio, we use the last prefill token as the query and compute the normalized attention assigned to dynamic tokens, $W_{\mathrm{dynamic}}$, at each layer. As shown in Fig.~\ref{fig:5} (right), FlashGaze consistently allocates more attention to dynamic tokens than the vanilla model, while improving accuracy from 71.6\% to 73.2\%, suggesting that allocating more attention to dynamic visual tokens may contribute to better video understanding.

\paragraph{Visualization of pruned videos.}
We visualize FlashGaze pruning on three video clips. In Fig.~\ref{fig:2}~(a) and (b), most backgrounds are static, and FlashGaze mainly retains regions that change over time, achieving retained ratios of only 10\% and 13\%. In contrast, the continuously changing driving scene in Fig.~\ref{fig:2}~(c) requires more patches. FlashGaze uses coarse-scale patches for low-information areas such as the sky, while preserving fine-scale patches for regions rich in detail such as traffic signs.

\section{Conclusion}
Long and high-resolution videos produce lengthy visual token sequences that impose substantial computational overhead on MLLMs, yet much of this cost stems from spatiotemporally redundant content. We propose FlashGaze, a training-free method that prunes redundant visual patches directly at the video input stage, before ViT encoding. Based on the observation that such redundancy can be assessed in pixel space, FlashGaze estimates the costs of dropping, merging, and keeping patches, and applies Quadtree DP to determine the action assignment under a given budget. Experiments on two representative MLLMs and six benchmarks show that FlashGaze retains 98\% of the full-input accuracy while achieving up to $5.4\times$ and $17\times$ speedups in ViT encoding and MLLM prefill, with $1.8\times$ lower peak GPU memory. FlashGaze further scales MLLMs to 1.5K-frame, 4K-resolution videos, unlocking longer and higher-resolution visual contexts for improved video understanding.

\newpage

\subsection*{AI use statement}
In this work, we used generative AI tools for retrieval and discovery, including literature search, related-work identification, and summarization of relevant papers. We also used generative AI tools to aid and polish writing by editing the manuscript for grammar, clarity, conciseness, and readability.
All AI-assisted content was reviewed and verified by the authors, who take full responsibility for the final content of the paper.

\subsection*{Reproducibility statement}
We provide detailed descriptions of the experimental implementation in Sec.~\ref{subsect:Exp_setup} and ~\ref{app:exp}. We will also provide source code in the supplementary materials.

\bibliography{iclr2027_conference}
\bibliographystyle{iclr2027_conference}

\clearpage
%=================================================
\appendix

\section{Additional Details of Quadtree DP}
\label{app:a}

This section provides a more detailed description of the overall procedure of the Quadtree DP introduced in Sec.~\ref{para:quadtree}. Algorithm~\ref{alg:quadtree} takes as input the action costs estimated in Sec.~\ref{para:cost estimation}, the patch budget $\mathcal{B}$, the penalty coefficient $\lambda$, and the current frame together with its corresponding quadtree $\mathcal{Q}$, and outputs the final action assignment $\mathcal{A}_t^{*}$. The quadtree structure is illustrated in Fig.~\ref{fig:3} (right). During DP, $\mathrm{Trace}$ records the minimizing child-count allocations and the decisions between merging a region and representing it through its children. These records are used for backtracking; their updates are omitted from the pseudocode for brevity.

\begin{algorithm}[H]
\caption{Quadtree DP}
\label{alg:quadtree}
\begin{algorithmic}[1]
\Require Quadtree $\mathcal{Q}$, action costs,
         patch budget $\mathcal{B}$, penalty $\lambda$,
         current frame $F_t$
\Ensure Action assignment $\mathcal{A}_t^{*}$

\ForAll{fine-scale leaf $p_{\mathrm{f}}$ in $\mathcal{Q}$}
    \State $T_{\mathrm{f}}(p_{\mathrm{f}},0)
           =\operatorname{Cost}(p_{\mathrm{f}},\mathrm{Drop})$,
           $T_{\mathrm{f}}(p_{\mathrm{f}},1)=0$
\EndFor

\ForAll{mid-scale node $p_{\mathrm{m}}$ in $\mathcal{Q}$
        with children $p_0,\ldots,p_3$}
    \State Apply \textsc{SCM} to
           $T_{\mathrm{f}}(p_0,\cdot),\ldots,
            T_{\mathrm{f}}(p_3,\cdot)$
           to obtain
           $\widetilde{T}_{\mathrm{m}}(p_{\mathrm{m}},\cdot)$
    \For{$n=0,\ldots,4$}
        \State $T_{\mathrm{m}}(p_{\mathrm{m}},n)
               =\widetilde{T}_{\mathrm{m}}(p_{\mathrm{m}},n)$
    \EndFor
    \State $T_{\mathrm{m}}(p_{\mathrm{m}},1)
           =\min\bigl(
           T_{\mathrm{m}}(p_{\mathrm{m}},1),
           \operatorname{Cost}(p_{\mathrm{m}},\mathrm{Merge})
           \bigr)$
\EndFor

\State Repeat the mid-scale procedure for all coarse-scale
       nodes $p_{\mathrm{c}}$ in $\mathcal{Q}$, using the
       mid-scale child cost arrays, $n=0,\ldots,16$, and
       $\operatorname{Cost}(p_{\mathrm{c}},\mathrm{Merge})$
       to obtain $T_{\mathrm{c}}(p_{\mathrm{c}},\cdot)$.

\State Apply \textsc{SCM} to the coarse-scale cost arrays
       until obtaining the frame-level array
       $T_{\mathrm{frame}}(\cdot)$.

\State $N_t^{*}
       =\displaystyle\arg\min_{N_t\in\{0,\ldots,\mathcal{B}\}}
       \bigl\{T_{\mathrm{frame}}(N_t)+\lambda N_t\bigr\}$

\State \Return
       \Call{Backtrack}{$\mathcal{Q},\mathrm{Trace},N_t^{*}$}
\end{algorithmic}
\end{algorithm}

\section{Additional Details of Integration with MLLMs}
\label{app:integ}
This section provides further implementation details on integrating FlashGaze with different MLLMs. For the Qwen3-VL series, the ViT natively supports multi-scale image inputs, allowing FlashGaze to be seamlessly integrated without any downstream fine-tuning. For NVILA-8B-Video, we enable multi-scale patch inputs by interpolating each frame and its positional embeddings to different scales, allowing patches at different scales to be processed by the ViT encoder.

%===========================================================
\section{Additional Details of Experiment Setup}
\label{app:exp}

All latency measurements are conducted on a single NVIDIA H200 GPU with a batch size of 1 and BF16 precision. We consistently use PyTorch native SDPA for attention instead of FlashAttention to ensure a fair comparison across methods. Each configuration is first warmed up with one run, followed by three consecutive runs whose latency is averaged. Peak GPU memory refers to the maximum GPU memory actually allocated during inference. We follow the official prompt protocols of each benchmark as implemented in lmms-eval.

%===========================================================

\section{Additional Details of Efficiency under Different Retained Ratios}
\label{app:b}

This section provides a detailed breakdown of the TTFT reported in Table~\ref{tab:3}. We divide TTFT into three stages: pre-ViT, which includes video processing and pruning overhead; ViT encoding; and MLLM prefill. Stage-wise latency results are reported below.

\begin{table*}[htb]
\caption{\textbf{Stage-wise latency under different retained ratios.}}
\centering
\small
\renewcommand{\arraystretch}{1.0}
\resizebox{\textwidth}{!}{%
\begin{tabular}{@{}c@{\hspace{12pt}}c@{}}
\setlength{\tabcolsep}{3pt}
% 左表
\begin{tabular}{lcccc}
\multicolumn{5}{c}{Latency on LongVideoBench} \\[2pt]
\toprule
\multirow{2}{*}{Method} &
Retained &
Pre-ViT &
ViT encoding &
MLLM prefill \\
& Ratio & (s) & (s) & (s) \\
\midrule
Qwen3-VL-8B & 100\% & 0.7 & 5.9 & 65.1\\
\midrule
%=================LVB-50%keep====================
+ToMe      & \multirow{4}{*}{50\%} & 0.4 & 5.9 & 18.0\\
+FlashVID  &                       & 2.2 & 6.6 & 17.7\\
+AutoGaze  &                       & 14.6 & 3.6 & 16.7\\
+FlashGaze&&0.5&2.7&17.5
\\
\midrule
%=================LVB-40%keep====================
+ToMe      & \multirow{4}{*}{40\%} & 0.6 & 5.9 & 12.1 \\
+FlashVID  &                       & 2.1 & 6.7& 11.9 \\
+AutoGaze  &                       & 10.7 & 2.1 & 12.5 \\
+FlashGaze&&0.5&2.1&11.9 \\
\midrule
%=================LVB-30%keep====================
+ToMe      & \multirow{4}{*}{30\%} & 0.5 & 5.9 & 7.4\\
+FlashVID  &                       & 2.0 & 6.7 & 7.3\\
+AutoGaze  &                       & 8.9 & 1.8 & 7.5\\
+FlashGaze&&0.5&1.6&7.4
\\
\midrule
%=================LVB-20%keep====================
+ToMe      & \multirow{4}{*}{20\%} & 0.5 & 5.8 & 4.0 \\
+FlashVID  &                       & 1.8 & 6.6 & 3.8 \\
+AutoGaze  &                       & 7.3 & 1.1 & 3.7 \\
+FlashGaze&& 0.5 &1.1&3.8
\\
\bottomrule
\end{tabular}
&
% 右表
\begin{tabular}{lcccc}
\multicolumn{5}{c}{Latency on HLVid} \\[2pt]
\toprule
\multirow{2}{*}{Method} &
Retained &
Pre-ViT &
ViT encoding &
MLLM prefill \\
& Ratio & (s) & (s) & (s) \\
\midrule
Qwen3-VL-8B & 100\% & 0.5 & 7.9 & 63.9\\
\midrule
%=================HLVID-50%keep====================
+ToMe      & \multirow{4}{*}{50\%} & 0.4 & 7.8 &17.5\\
+FlashVID  &                       & 2.2 & 9.5 & 17.2\\
+AutoGaze  &                       & 14.6 & 3.1&16.9\\
+FlashGaze&&0.5&3.0&17.0
\\
\midrule
%=================HLVID-40%keep====================
+ToMe      & \multirow{4}{*}{40\%} & 0.5 & 7.8 &11.7\\
+FlashVID  &                       & 2.0&9.4 &11.6 \\
+AutoGaze  &                       & 11.3 & 2.4 & 11.7\\
+FlashGaze&&0.5&2.5&11.3
\\
\midrule
%=================HLVID-30%keep====================
+ToMe      & \multirow{4}{*}{30\%} & 0.5 & 7.8 &7.1\\
+FlashVID  &                       & 1.7 & 9.5 &7.0\\
+AutoGaze  &                       & 10.0 & 1.7&7.3\\
+FlashGaze&&0.5&1.6&7.0
\\
\midrule
%=================HLVID-20%keep====================
+ToMe      & \multirow{4}{*}{20\%} & 0.6 & 7.7 & 3.8\\
+FlashVID  &                       & 1.6 & 9.6 & 3.6\\
+AutoGaze  &                       & 8.3 & 1.5 & 3.7\\
+FlashGaze&& 0.5&1.1&3.7
\\
\bottomrule
\end{tabular}
\end{tabular}%
}
\label{tab:lat_diff_stage}
\end{table*}

We define the retained ratio as the number of visual tokens remaining after the final pruning stage divided by the original number of visual tokens. For FlashVID, the ViT-stage pruning reduces the tokens only to an intermediate ratio, while a subsequent pruning stage further reduces them to the final retained ratio. Therefore, the ViT processes the less-pruned intermediate sequence, resulting in higher ViT latency. AutoGaze has higher pre-ViT latency because it uses an autoregressive network to process the video before ViT encoding. In contrast, FlashGaze performs pruning only once, before the ViT, without introducing additional networks. Thus, FlashGaze achieves the lowest overall latency.

%==========================================================

\section{Additional Details of Analysis}
\label{app:c}

This section describes the detailed computation of visual feature similarity. Since the vanilla input and FlashGaze may contain different numbers of visual tokens for the same spatial area after pruning, their ViT output features cannot be directly compared. We therefore aggregate the visual features using the attention weights from the first MLLM layer to obtain a fixed-dimensional representation for comparison. Let $\mathcal{I}'_t$ denote the indices of the visual tokens corresponding to the pruned frame $F'_t$. Using the last prefill token as the query, we compute the representative feature as
\begin{equation}
\mathbf{x}_t
=
\mathbf{W}_{\mathrm{O}}\,
\operatorname{Concat}_{h}
\left(
\sum_{i\in\mathcal{I}'_t}
\alpha_{h,i}\mathbf{v}_{h,i}
\right).
\end{equation}
Here, $h$ indexes attention heads, $i$ indexes visual tokens,
$\alpha_{h,i}$ is the attention weight assigned to token $i$
by head $h$, $\mathbf{v}_{h,i}$ is the corresponding value
vector, and $\mathbf{W}_{\mathrm{O}}$ is the output
projection matrix.

For the vanilla input, we identify the corresponding spatial
area in the original frame and apply the same aggregation
to all visual tokens within that area, obtaining $\mathbf{y}_t$.
We then compute their cosine similarity as
\begin{equation}
\operatorname{Similarity}(\mathbf{x}_t,\mathbf{y}_t)
=
\frac{\mathbf{x}_t^\top\mathbf{y}_t}
{\lVert\mathbf{x}_t\rVert_2\,\lVert\mathbf{y}_t\rVert_2}.
\end{equation}

\section{Additional Visual Cases of Pruned Videos}
\label{app:d}
In this section, we present more video clips processed by FlashGaze.
\begin{figure}[H]
    \centering
    \includegraphics[width=\linewidth]{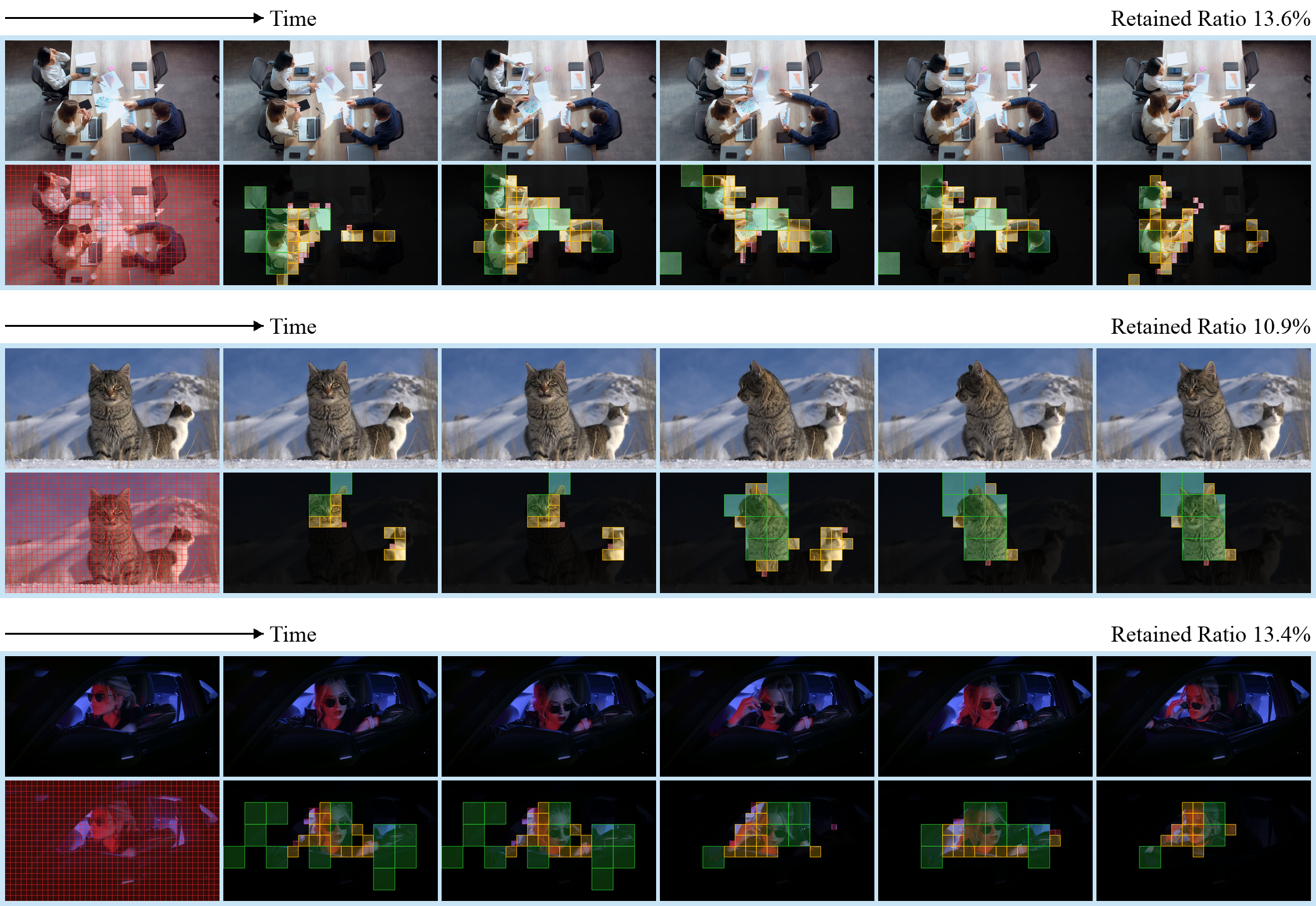}
    \caption{\textbf{More cases of videos pruned by FlashGaze.} }
    \label{fig:appdx}
\end{figure}

\end{document}